\documentclass[10pt,twocolumn,letterpaper]{article}

\usepackage{cvpr}
\usepackage{times}
\usepackage{epsfig}
\usepackage{graphicx}
\usepackage{amsmath}
\usepackage{amssymb}
\usepackage{microtype}
\usepackage{caption}
\usepackage{subcaption}
\usepackage{balance}

\usepackage[numbers, sort, round]{natbib}
\setcitestyle{comma}

\usepackage[pagebackref=true,breaklinks=true,letterpaper=true,colorlinks,bookmarks=false]{hyperref}

\cvprfinalcopy 

\def\cvprPaperID{5} 

\begin{document}

\title{NDD20: A large-scale few-shot dolphin dataset for coarse and fine-grained categorisation}

\newcommand*{\affmark}[1][*]{\textsuperscript{#1}}
\newcommand*{\email}[1]{\texttt{#1}}

\author{%
Cameron Trotter \affmark[2]\and Georgia Atkinson \affmark[2]\and Matt Sharpe \affmark[3]\and Kirsten Richardson \affmark[1]\and A. Stephen McGough \affmark[1]\and Nick Wright \affmark[2]\and Ben Burville \affmark[3]\and Per Berggren \affmark[3]\and\\
\affmark[1] School of Computing\\
\affmark[2] School of Engineering\\
\affmark[3] School of Natural and Environmental Sciences\\
Newcastle University, Newcastle Upon Tyne, UK\\
\tt\small \{c.trotter2, g.atkinson, m.j.sharpe, k.crane2,\\ 
\tt\small stephen.mcgough, nick.wright, ben.burville, per.berggren\}@ncl.ac.uk\\
}

\maketitle
\begin{abstract}
   We introduce the Northumberland Dolphin Dataset 2020 (NDD20), a challenging image dataset annotated for both coarse and fine-grained instance segmentation and categorisation. This dataset, the first release of the NDD \cite{trotter2019northumberland}, was created in response to the rapid expansion of computer vision into conservation research and the production of field-deployable systems suited to extreme environmental conditions - an area with few open source datasets. NDD20 contains a large collection of above and below water images of two different dolphin species for traditional coarse and fine-grained segmentation. All data contained in NDD20 was obtained via manual collection in the North Sea around the Northumberland coastline, UK. We present experimentation using standard deep learning network architecture trained using NDD20 and report baselines results.
\end{abstract}

\section{Introduction}\label{sec:Introduction}

Conservation is an area with great potential for computer vision utilisation. Cetacean conservation (dolphins, whales and porpoises) could especially benefit from the introduction of computer vision aides; when studying cetacean population dynamics and health for example, researchers often focus on methods such as photo-id - identifying individuals based on unique characteristics. This manual identification process often takes many months to complete, and thus any help from computer vision systems capable of fine-grained cetacean classification would afford researchers more time in the field and less time processing collected data.

Very few open-source datasets exist for use within a conservation or ecological space; those that do often focus on simple object detection of animals in a scene. Datasets often focus on a particular subset of animals such as pets commonly found in homes \cite{parkhi12a, khosla2011novel} or birds \cite{wah2011caltech, van_horn_building_2015}. Some large scale datasets showing animals in natural environments do exist \cite{Horn_2018_CVPR}, although these often only provide labels at a species level, which is not fine-grained enough for population estimation which requires the identification of individuals. Of the datasets which do allow for species identification currently, most primarily focus on the development of land-based camera trap systems \cite{beery2019iwildcam, wilber2013animal}, although work has also been undertaken in the development of marine life species detection systems \cite{anwar2015invariant}.

This may seem like an almost impossible task, however it has been achieved by cetacean researchers for over forty years \cite{wursig1977photographic}, manually identifying individuals based on prominent markings on their fins as they breach the waterline. As such it is believed that an automated photo-id process is achievable using computer vision \cite{weinstein_computer_2018}, with work already beginning in this area \cite{gilman_computer-assisted_2016, bouma_individual_2018-1, weideman_integral_2017, hale_unsupervised_2012}.

Development of automatic photo-id is currently limited to niche research groups who have the domain expertise in cetacean research as well as the technical and mathematical ability to develop automated systems to utilise this. These research groups often build systems using their own private datasets based on cetacean catalogues in their geographical area, and thus it is hard to benchmark competing systems. NDD20 provides researchers with an open source dataset that is both true to the deployment environment and fine-grained to an individual level, thus allowing for photo-id benchmarking across systems for the first time, as well as allowing researchers interested in this area to develop novel systems without the need to perform their own data collection. 


\section{Data Collection}\label{sec:DataCollection}

Collection of the data found in NDD20 was achieved through two separate fieldwork efforts. Below water collection efforts consisted of 36 opportunistic surveys of the Farne Deeps - a glacial trench with high bio-productivity approximately 14-20nm offshore from the village of Seahouses, UK, from 2011-2018 \cite{vanbressem2018visual}. Above water efforts consisted of 27 surveys along a stretch of the Northumberland coast designated as the Coquet to St. Mary's Marine Conservation Zone (MCZ). 

Above water photographs were taken using a DSLR camera from the deck of a small rigid inflatable boat on days of fair weather and good sea conditions (less than four on the Beaufort scale). Below water images in the dataset are screen grabs from high-definition video footage taken with a GoPro Hero 3 and GoPro Hero 4 cameras, captured by a diver, again under good sea conditions. Above water surveys were initially conducted in a manner that followed predetermined transect lines, but later moved to more opportunistic surveying, with the use of shore-based volunteer observations posted on dedicated social media pages. 


\section{Properties of NDD20}\label{sec:Properties}

NDD20 contains a range of image data which is split into two main categories; above and below water. Above water images taken from the deck of a research vessel represent the standard data format utilised by cetacean researchers today, whereby individuals are identified using the structure of the dolphin's dorsal fin. Below water images are less common, but provide additional features for identification such as general colouring, unique body markings, scarring and patterns formed by injury or skin disease.

To protect ongoing cetacean research efforts a pseudoanonomisation has been performed. This, however, does not diminish the value of this data to computer vision researchers. It is not the case that images with sequential filenames were captured sequentially, and all individual IDs have been randomly allocated a numerical value rather than the code given to them by the Northumberland cetacean researchers. All EXIF data found in the images has also been removed.

\subsection{Above Water Images}\label{subsec:AboveWaterImages}
In total, NDD20 contains 2201 above water images. These images are accompanied by a JSON file in via-2.0.8 label format \cite{dutta2019vgg}. This JSON file contains, for each image, a set of (x,y) coordinates denoting regions of interest in the image. Each of these sets of coordinates is accompanied with attribute labels, denoting different levels of difficulty in the recognition task. The first attribute level is \texttt{dolphin} which labels the area of the image containing any part of a dolphin which was above the waterline when the image was captured, representing a standard (although difficult) instance segmentation task. In total, approx 2900 masks are present in the above water data, as some images contain more than one mask.

The second attribute is the dolphin's species, either \texttt{BND} or \texttt{WBD}, representing the species \textit{Tursiops truncatus} and \textit{Lagenorhynchus albirostris} respectively. These attributes provide a fine-grained categorisation challenge, as there are small inter-species differences between the two classes. All above water masks contain this label, although the distribution of the classes is imbalanced with 73\% being labelled \texttt{BND}. 

\begin{figure}
    	\begin{center}
    		\includegraphics[width=\linewidth]{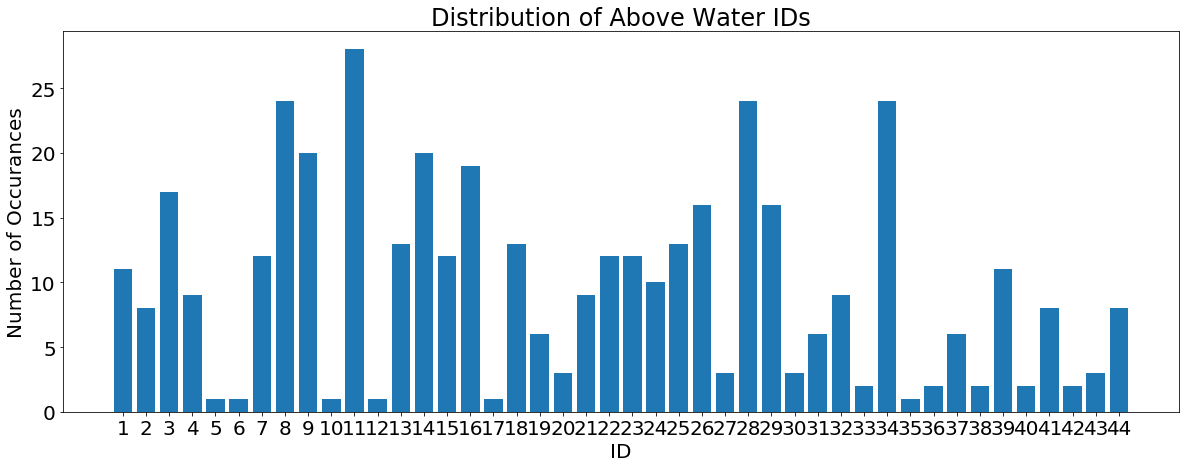}
    	\end{center}
    	\caption{The number of above water images per ID class.}
    	\label{fig:aboveWaterDist}
\end{figure}

The final attribute label represents the individual identifications of the dolphins. These individuals were manually identified by cetacean researchers specialising in the Northumberland coastal area. In total, around 14\% of masks contain an \texttt{ID} attribute with 44 unique classes present. Once again, these classes are imbalanced presenting both a fine-grained and few-shot learning problem. The number of above water images per ID class can be seen in Figure \ref{fig:aboveWaterDist}.

\begin{figure}
    	\begin{center}
    		\includegraphics[width=0.85\linewidth]{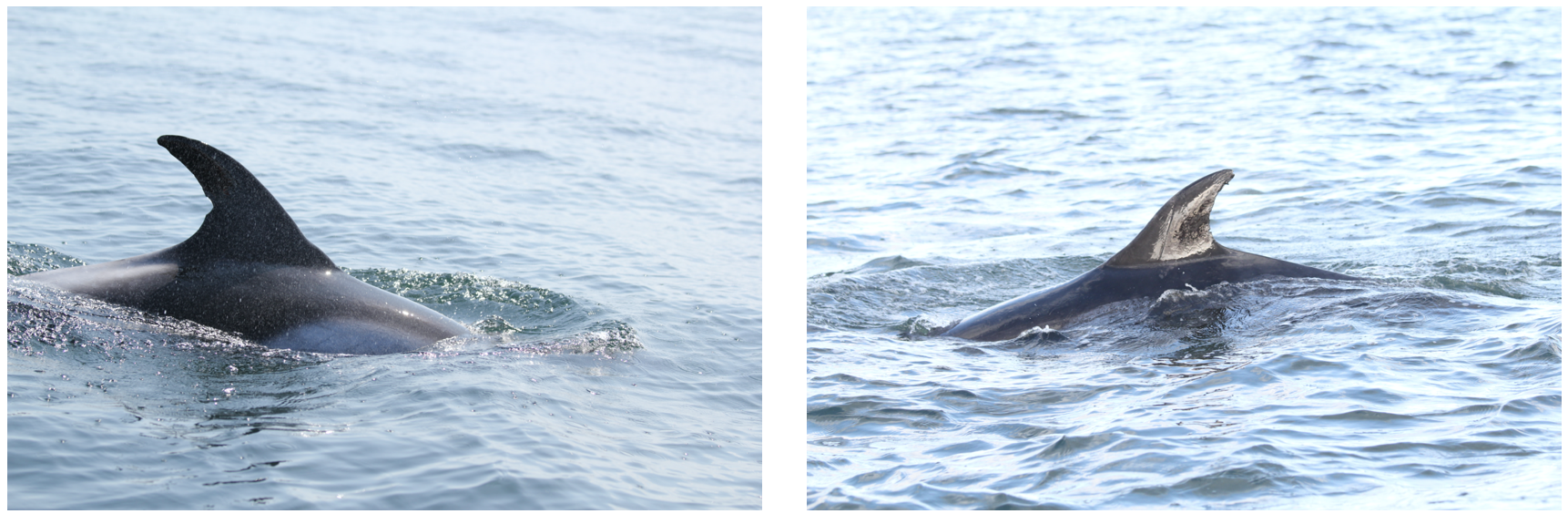}
    	\end{center}
    	\caption{Example above water images. Both images contain one mask with the following attributes: Left - \texttt{object:dolphin}, \texttt{species:WBD}, \texttt{id:11}. Right - \texttt{object:dolphin}, \texttt{species:BND}, \texttt{id:8}.}
    	\label{fig:aweg}
\end{figure}

Classification based on ID class labels represents the hardest fine-grained challenge in NDD20, as individual identification outside of human facial recognition is still a relatively unexplored research area. In order to identify individuals in the dataset, intra-class differences such as scarring or pigmentation on the surface of the individual must be taken into account, as well as the individual dorsal fin shape. Further to this, only a small number of examples for each individual exists resulting in both a fine-grained and a few shot learning task. Many challenges exist in the above water data which must be overcome. These include images with small regions of interest, large amounts of blur or water splash, the dolphin fin being partially obstructed or being photographed from an angle, and markings which can be utilised for identification only being present on one side of the fin. Example above water images can be seen in Figure \ref{fig:aweg}.

\subsection{Below Water Images}\label{subsec:BelowWaterImages}

\begin{figure}
    	\begin{center}
    		\includegraphics[width=\linewidth]{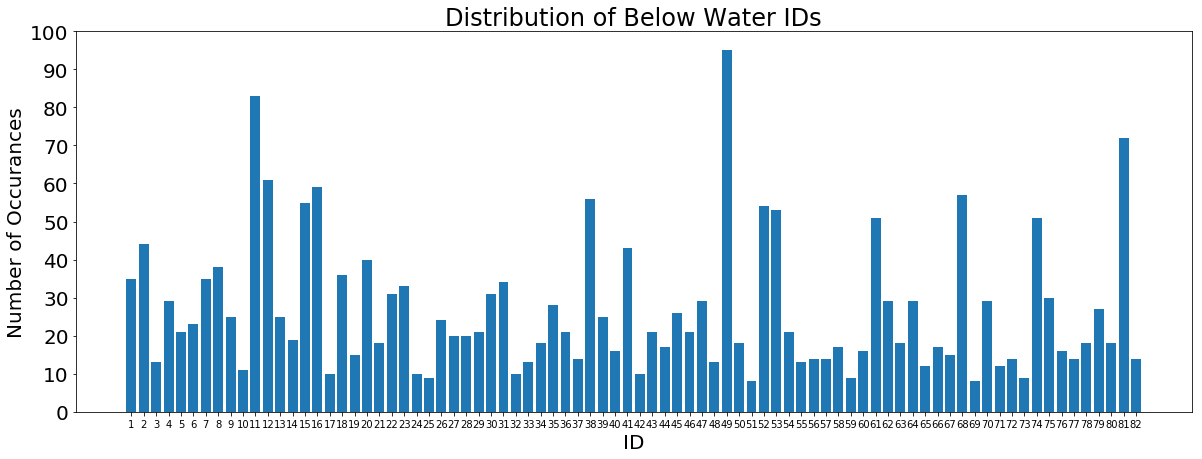}
    	\end{center}
    	\caption{The number of below water images per ID class.}
    	\label{fig:belowWaterDist}
\end{figure}

The 2201 below water images presented are a subset of a much larger collection of images collected as far back as 2011. The labelling of these images mirrors the above water images - we provide a JSON file containing the coordinates of a manually drawn mask annotation as well as multiple attribute labels. Again, the first attribute level is \texttt{dolphin}, this time corresponding to the area of the image containing any part of the object visible. Unlike the above water images, all below water images contain at least one mask with an \texttt{ID} attribute, with 82 classes. Some ID class labels appear significantly fewer times than others due to the nature of collection of data on free roaming individuals, presenting an interesting and true to life few-shot learning problem. The number of below water images per ID class can be seen in Figure \ref{fig:belowWaterDist}. No \texttt{species} label is provided as all images are of \textit{Lagenorhynchus albirostris}. Below water images are also labelled with an \texttt{out of focus} flag, denoting if the individual is deemed to be out of focus.

\begin{figure}[h]
    	\begin{center}
    		\includegraphics[width=\linewidth]{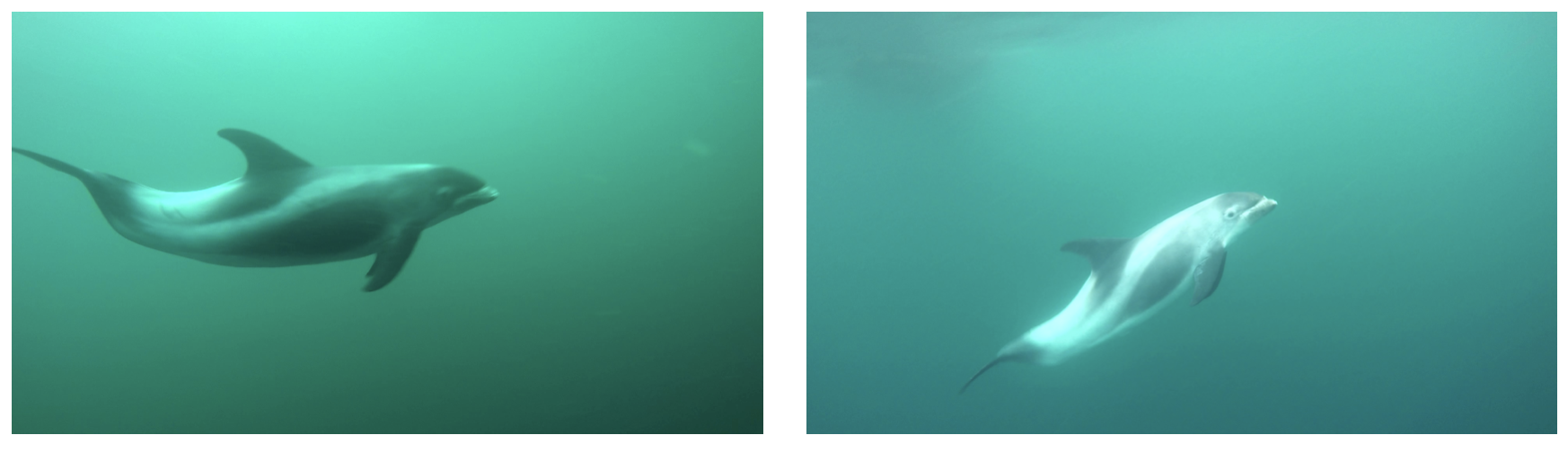}
    	\end{center}
    	\caption{Example below water images. Both images contain one mask with the following attributes: Left - \texttt{object:dolphin}, \texttt{id:9}, \texttt{out of focus:false}. Right - \texttt{object:dolphin}, \texttt{id:30}, \texttt{out of focus:false}.}
    	\label{fig:uweg}
\end{figure}

The main challenges with respect to the below water images are water clarity, affected by factors such as algae bloom, and sunlight refraction which may both obscure areas of the individual useful for identification or add artificial markings which may hinder this.  Many of the challenges associated with above water images apply here too, particularly the likelihood that unique features are specific to one side of the body. Examples of below water images in the dataset can be seen in Figure \ref{fig:uweg}.

\section{Baseline Experimentation}\label{sec:BaselineExperimentation}

\begin{figure}
    	\begin{center}
    		\includegraphics[width=\linewidth]{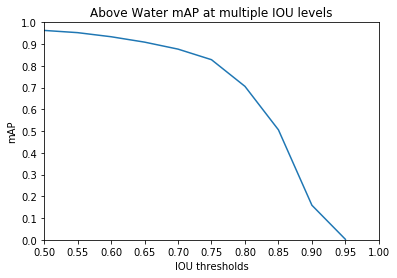}
    	\end{center}
    	\caption{Differing mAP@IOU values for the best performing instance segmentation model using the above water data.}
    	\label{fig:map}
\end{figure}

In order to demonstrate both the usefulness and the challenging nature of the dataset, the following baseline experimentation was performed. Using the above water data, a Mask-RCNN model was trained to perform instance segmentation utilising the \texttt{dolphin} mask label \cite{DBLP:journals/corr/HeGDG17}. The model, built using the ResNet-50 architecture \cite{he_deep_2015} pretrained on the MSCOCO dataset \cite{10.1007/978-3-319-10602-1_48} using a cosine annealing learning rate scheduler with periodic restarts \cite{loshchilov2016sgdr}, was trained for 100 epochs using a random 80:20 train-test split resulting in mAP@IOU[0.5, 0.75] = [0.96, 0.83], excellent results giving the difficulty of the dataset. The full range of mAP@IOU values can be seen in Figure \ref{fig:map}.

\begin{figure}
    	\begin{center}
    		\includegraphics[width=\linewidth]{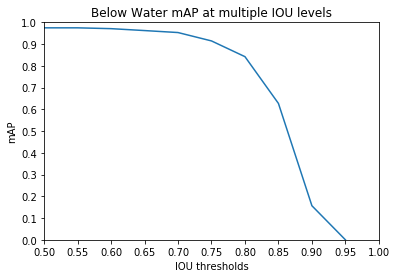}
    	\end{center}
    	\caption{Differing mAP@IOU values for the best performing instance segmentation model using the below water data.}
    	\label{fig:BWmap}
\end{figure}

This instance segmentation task was also performed using the below water data. Here, a second Mask-RCNN model was trained and achieved mAP@IOU[0.5, 0.75] = [0.97, 0.91]. These results show the baseline model trained on the below water data is not as accurate as the model trained with the above water data. This may be due to the larger variation in the shape of masks in the below water data as a result of the underwater cameras being able to capture the individual cetaceans in a wider range of movement, as well as there being less differentiation between the background and the individual due to water conditions. The full range of mAP@IOU values can be seen in Figure \ref{fig:BWmap}.

\section{Conclusion}\label{sec:Conclusion}

With a large quantity of self-collected data and a multidisciplinary team, it has been possible to leverage domain-specific knowledge in order to develop a varied dataset of images with multi-tiered annotations. Quality datasets of this sort are instrumental for developing high-accuracy computer vision algorithms. Not only is this progress key to the field of computer science, in this particular instance it is also key to marine science, providing a valuable tool for automated photo-id that will afford researchers more time in the field and less time performing time-consuming, manual labelling. 

This first version release of NDD20 provides the user with a large collection of above and below water images that have been annotated with a hierarchy of levels. Since data collection of the kind described above is continuing, it is likely that future releases of the dataset will contain a growing number of images. 

To demonstrate the usefulness of the dataset and inspire future work, we report benchmark results using the above and below water images for an object segmentation task. This experimentation was purely to provide an idea of baseline performance, and future work will focus on producing baseline results for other tasks achievable with NDD20 such as fine-grained species and individual classification. It is our hope and expectation that the dataset be used to attain higher levels of accuracy than those seen in the segmentation baseline, to perform categorisation at the level species and individual, and to perhaps even pre-train models that will generalise to other animals or objects. 

\begin{flushleft}\textbf{NDD20 can be accessed at: \href{https://doi.org/10.25405/data.ncl.c.4982342}{https://doi.org/10.25405/data.ncl.c.4982342}}\end{flushleft}

\section{Acknowledgements}\label{sec:Acks}

This work was supported by the Engineering and Physical Sciences Research Council, Centre for Doctoral Training in Cloud Computing for Big Data [grant number EP/L015358/1], as well as the Ridley Fellowship awarded by Newcastle University.

{
\bibliographystyle{ieee_fullname}
\balance
\bibliography{references}
}

\end{document}